# Steel Plate Fault Detection using the Fitness Dependent Optimizer and Neural Networks

Salar Farahmand-Tabar[1*] · Tarik A. Rashid [2]

[1]Department of Civil Engineering Eng., Faculty of Engineering, University of Zanjan, Zanjan, Iran,

[2]Department of Computer Science and Engineering, University of Kurdistan Hewler, Erbil, KR, Iraq

*farahmandsalar@znu.ac.ir; farahmandsalar@gmail.com
 tarik.ahmed@ukh.edu.krd

**Abstract.** Detecting faults in steel plates is crucial for ensuring the safety and reliability of the structures and industrial equipment. Early detection of faults can prevent further damage and costly repairs. This chapter aims at diagnosing and predicting the likelihood of steel plates developing faults using experimental text data. Various machine learning methods such as GWO-based and FDO-based MLP and CMLP are tested to classify steel plates as either faulty or non-faulty. The experiments produced promising results for all models, with similar accuracy and performance. However, the FDO-based MLP and CMLP models consistently achieved the best results, with 100% accuracy in all tested datasets. The other models' outcomes varied from one experiment to another. The findings indicate that models that employed the FDO as a learning algorithm had the potential to achieve higher accuracy with a little longer runtime compared to other algorithms. In conclusion, early detection of faults in steel plates is critical for maintaining safety and reliability, and machine learning techniques can help predict and diagnose these faults accurately.

## 1. Introduction

Surface defects on steel products can have a significant impact on their quality [1]. Not only do these defects impact the later stages of production, but they also undermine the ability of the final products to withstand corrosion and wear. In order to identify these flaws, inspection systems employ CCD cameras to capture images of the steel surface under specific lighting conditions. Subsequently, defect identification algorithms are utilized to analyze the images and identify and categorize the surface defects. However, designing algorithms for the detection and classification of surface defects has proven to be a challenging task due to their rarity and variations in appearance. As illustrated in Fig. 1, steel plates commonly exhibit two types of surface defects: seams (a)-(d) and scales (e)-(h). The surface defects found on steel plates show considerable differences within each category, and the image background is often intricate.

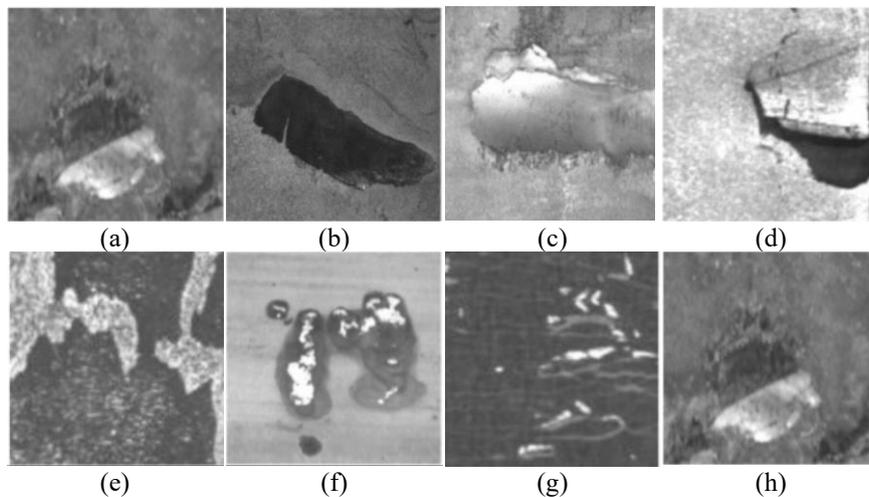

(a)  (b)  (c)  (d)
(e)  (f)  (g)  (h)

**Fig. 1.** Types of typical defects on hot rolled plates: 1) seams [(a)–(d)] and 2) scales [(e)–(h)] [2]

There are four distinct approaches to detect surface defects, which include statistical, model-based, structural, and filter-based techniques. The statistical approach utilizes properties such as histogram, autocorrelation, local binary patterns, and Gray Level Co-occurrence Matrix (GLCM). GLCM can determine various statistics of texture, such as entropy, dissimilarity, energy, correlation, homogeneity, and contrast. The model-based approach involves creating mathematical models to represent surface defects and comparing the actual surface with the models to detect deviations. The structural approach involves examining the geometric properties of the surface, such as edges and contours, to detect defects. The filter-based approach uses various filters to enhance the appearance of surface defects.

Despite the availability of various approaches to surface defect detection, there are still some limitations that need to be addressed. One significant challenge is the need for specialized illumination and imaging equipment, which can be expensive and difficult to maintain. Additionally, the accuracy of detection algorithms can be affected by changes in lighting conditions and surface geometry. As such, there is a need to continue developing more robust and adaptable algorithms that can overcome these limitations and provide accurate and reliable detection of surface defects.

In recent years, there have been significant advancements in algorithms designed to detect and classify defects on steel surfaces. For instance, a method utilized an extreme learning machine (ELM) and a well-known genetic algorithm (GA) [3] to classify hot-rolled plate defects [4]. By employing this method, the durability, and effectiveness of the ELM in inspecting steel surfaces were improved. Using enhanced metaheuristics [5, 6] can make further improvements in the process. Another strategy involved incorporating the RNAMlet into surface inspection as a feature extractor, enabling the image to be decomposed asymmetrically [7]. As a result, the adaptability of the feature extraction process was enhanced, enabling its application across various steel production lines.

Furthermore, surface inspection algorithms have been devised utilizing the scale-invariant feature transform (SIFT) and support vector machine (SVM) [8]. These algorithms have demonstrated favorable detection outcomes in production lines where image backgrounds are uncomplicated and uncluttered. Another notable advancement in surface inspection involves the utilization of the shearlet transform (ST) to offer an effective representation of defects at multiple scales [9]. By employing this technique, the precision of defect recognition on steel surfaces with intricate backgrounds was significantly enhanced.

In the detection of surface cracks on structural steels, a combination of discrete Fourier transform (DFT) and artificial neural network was employed [10]. Additionally, Gabor filters were utilized to identify corner and thin cracks in raw steel blocks by minimizing the cost function that separates the energy characteristics of defective and defect-free regions [11]. Furthermore, an online system for crack detection was designed, which relied on 3D contour data from the steel plate surface and integrated image processing with statistical classification using logistic regression [12]. These various algorithms exemplify the ongoing progress and advancement in techniques for detecting and categorizing defects on steel surfaces.

Hence, surface inspection algorithms need to possess the ability to classify intricate defects and learn from data that lacks labels. In recent times, deep learning techniques have displayed exceptional performance in tasks like image classification and object detection. One widely used deep learning method is the Convolutional Neural Network (CNN), which was initially introduced by LeCun [13] and has proven to be highly effective in image classification. Unlike other classification algorithms, CNN does not rely on explicit feature extraction processes; instead, it learns the weights of the convolutional layers by minimizing the loss function. This empowers CNN to achieve superior classification results even when confronted with images featuring complex backgrounds and variations in appearance. Additionally, several variations of CNN have been proposed, such as Alexnet [14], VGG [15], NIN [16], Inceptions [17], Inception-Resnet [18], and Densenet [19]. These variants often incorporate deeper stacked convolutional layers or employ asymmetrical structures to extract more nonlinear features, thereby enhancing the CNN's performance in processing complex images. Furthermore, recent studies have demonstrated the effectiveness of CNN in tackling more demanding tasks, including object detection [20-26].

However, the effectiveness of CNN-based approaches heavily depends on the availability of an adequate number of training samples. When training a CNN with small datasets, the algorithm's ability to generalize can be significantly affected, thereby limiting its applicability in industrial settings. Presently, the most viable solution for utilizing CNN with small datasets is transfer learning [27]. This method works based on the assumption that the sample images within our specific field exhibit basic characteristics, such as edges and curves, which are also present in the images used to train existing models. Transfer learning can be employed in conjunction with convolutional neural networks (CNN) to train smaller datasets for tasks like emotion recognition [27] and automated medical diagnosis [28-30]. However, when it comes to steel surface inspection, the effectiveness of applying transfer learning is not as pronounced as in other domains. This limitation primarily stems from the substantial disparity in image context between steel surfaces and the majority of pre-existing models, thereby violating the application requirements of transfer learning.

Considering the previous works, it is clear that machine-learning techniques have shown promising results in diagnosing and predicting faults in steel plates. In this chapter, the use of various machine learning algorithms, such as MLP and CMLP optimized with GWO, MGWO, and FDO has demonstrated the potential for accurately classifying steel plates as either faulty or non-faulty considering a dataset. While all the models produced similar accuracy results, FDO_MLP and FDO_CMLP consistently achieved 100% accuracy in all tested datasets. This suggests that using the FDO algorithm as a learning algorithm may lead to higher accuracy with a slightly longer runtime compared to other algorithms. These findings highlight the importance of early detection of faults in steel plates for maintaining safety and reliability and the significant role that machine learning techniques can play in achieving this goal.

## 2. Research Methodology

This study involves the comparison of several different approaches. Specifically, five models with distinct network architectures or training algorithms were utilized and assessed across the dataset. The methodology of this investigation involves the identification of each dataset and the preparation of its data, the selection of appropriate features, the application of the classification model, and the utilization of statistical techniques to draw comparisons between the outcomes of each method (Fig. 2).

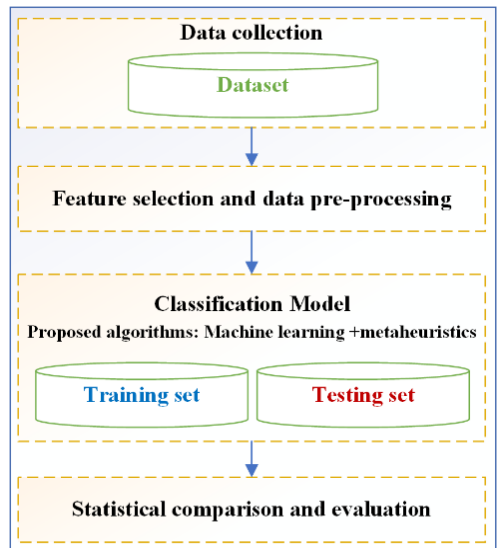

**Fig. 2.** Methodology of the research

### 2.1. Data Collection and Feature Selection

The crucial initial stage in machine learning is constructing a database. The utilized dataset is a unique steel plate fault dataset that originated in the research center of Sciences of Communication, Italy [31]. It comprises steel plate faults, classified into 6 different types Pastry, Z-Scratch, K-Scratch, Stains, Dirtiness, and Bumps. There are 29 variables including Min. and Max. of X, Min. and Max of Y, Pixels Areas, X and Y Perimeter, Sum of Luminosity, Min. and Max. of Luminosity, Length of Conveyer, TypeOfSteel_A300 and A400, Steel Plate Thickness, Edges Index, Empty Index, Square Index, Outside-X Index, Edges-X Index, Edges-Y Index, Outside-Global Index, Log. of Areas, Log. X Index, Log. Y Index,

Orientation Index, Luminosity Index, Sigmoid of Areas. The aim was to train machine learning for automatic pattern recognition [32, 33].

To preprocess the data for model training, the positive and negative input features are encoded as 0 and 1. In terms of the target, we used 1 to represent "positive" and 2 to represent "negative".

### 2.2. Model of Classification

This section outlines the methodology of the considered models, which includes the model architecture and the training methods explained in the following sections.

### 2.2.1. The architecture of the neural network

A neural network's architecture comprises three crucial components: the overall count of layers, encompassing both the input and output layers, the existence and quantity of hidden layers, and the number of nodes within each layer. The process of selecting an optimal topology for a neural network involves regulating the number of hidden layers and neurons within each layer [34]. In all of our proposed models, we utilized a solitary hidden layer, and the number of neurons in that layer was determined by the number of features present in the dataset. The criterion for determining the number of neurons in the hidden layer is as follows:

$$H_{no} = 2 * I_{no} + 1 \qquad (1)$$

Within this context, we utilize the symbols $H_{no}$ and $I_{no}$, representing the number of hidden layers and input layers respectively. Our study incorporates two types of neural networks: basic feed-forward artificial neural networks and cascade feed-forward artificial neural networks. The initial arrangement comprises three interconnected layers: an input layer, a hidden layer, and an output layer. In this configuration, the artificial neural network (ANN) establishes connections from the input layer to each subsequent layer and between each layer and the subsequent ones. On the other hand, the cascade ANN introduces an additional connection that directly links the input layer to the output layer. This allows the network to learn intricate associations and enhances the speed at which it acquires the desired relationship [35, 36].

### 2.2.2. Learning Method of Artificial Neural Network

The primary goal during training is to determine the most effective weights and biases that yield the highest possible accuracy in classification. In this study, various models utilize either GWO (Grey Wolf Optimizer), modified GWO, or FDO (Fitness Dependent Optimizer) as their training algorithms. The optimization process revolves around adjusting the weights and biases, to minimize an associated objective function. In neural networks, the Mean Square Error (MSE) serves as a commonly adopted evaluation metric. It quantifies the average squared difference between the desired output and the output generated by the ANN model. The MSE can be calculated using the following formula:

$$MSE = \sum_{i=1}^{n}(y_i - \hat{y}_i)^2 \qquad (2)$$

The calculation of the Mean Square Error (MSE) is given by the equation where $i$ is the input unit iteration, $n$ is the number of outputs, $y_i$ represents the desirable output, and $\hat{y}_i$ denotes the achieved value. To assess the model's performance, we compute the MSE for all training samples and then average the results. The optimization algorithm utilizes the computed average MSE value to make adjustments to the weights and biases of the model, aiming to minimize the average MSE across all training samples. The average MSE is calculated using the formula provided below:

$$MSE_{avg} = \sum_{j=1}^{m}\sum_{i=1}^{n}\frac{(y_i - \hat{y}_i)^2}{m} \qquad (3)$$

The application of the optimization algorithm to update the weights of neural network and achieve maximum accuracy is illustrated in Fig. 3.

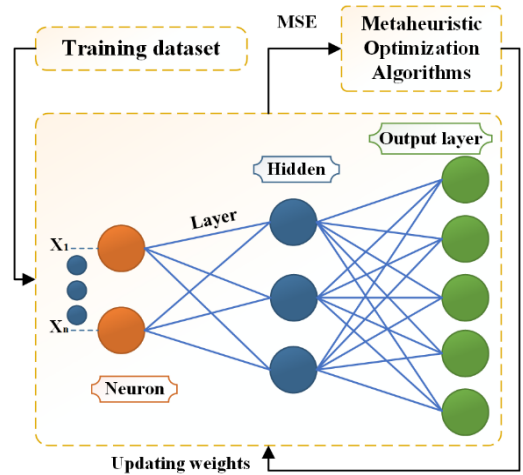

**Fig. 3.** Training procedures of the proposed models.

## 2.3. Result Evaluation

As mentioned earlier, the evaluation of the applied models involves two approaches: the Mean Square Error (MSE) value and the confusion matrix. Models that yield a very low MSE value, approaching zero, are regarded as effective. The confusion matrix provides a summary of the accurate and inaccurate predictions made by the models. Sensitivity (true positive rate), specificity (true negative rate), Positive Predictive Value (PPV), and Negative Predictive Value (NPV) are metrics obtained from the confusion matrix to understand the model's reliability. These metrics offer a measure of the probability that a sample classified as positive is indeed positive, or that a sample classified as negative is truly negative. Their values lie between 0 and 1, where a value closer to 1 signifies a more favorable result. A value of 1 represents the optimal outcome, while a value of 0 indicates the poorest performance. [37].

## 3. Optimization Algorithms for Weight and Bias Updating

In this research, three optimization algorithms were employed: the standard and Modified Grey Wolf Optimizer, and Fitness Dependent Optimizer. These algorithms were utilized to find the most suitable weight and bias values for a neural network used for predicting the presence or absence of faults in steel plates. The specifics of each algorithm will be discussed in this section.

## 3.1. Grey Wolf Optimization

Grey Wolf Optimization (GWO) is a type of swarm intelligence algorithm that was introduced by Mirjalili et al. [38]. This algorithm is based on the hunting behavior of grey wolves in nature. Grey wolves are recognized for their adeptness in efficiently hunting prey, employing a distinct approach. The GWO algorithm imitates this behavior by simulating the organization of tasks within a wolf pack, which is characterized by a social hierarchy. This social hierarchy represents the optimality of solutions, with the alpha ($\alpha$) being the most preferred solution, followed by the beta ($\beta$) and delta ($\delta$), and finally the other solutions are classified as omega ($\omega$). During a hunt, the grey wolf will first encircle the prey. The position of the grey wolf ($\overrightarrow{X_p}(t)$) around the prey ($\overrightarrow{X}(t)$) is updated using Eq. (4).

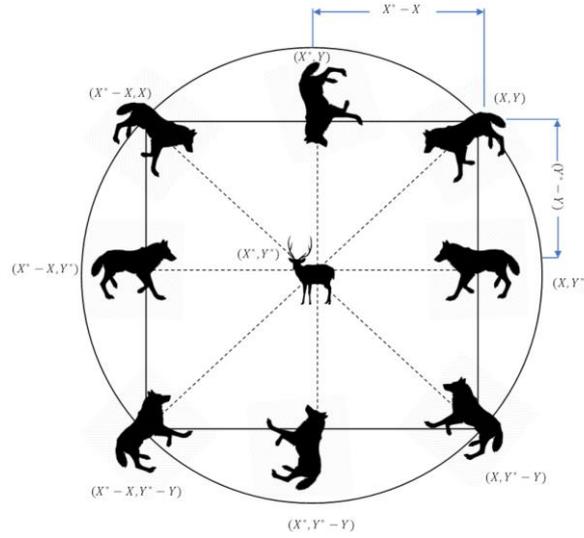

**Fig. 4.** Position vectors in two-dimensional space.

$$\overrightarrow{D} = \left| \overrightarrow{C} \cdot \overrightarrow{X_p}(t) - \overrightarrow{X}(t) \right| \tag{4}$$

$$\vec{x}(t+1) = \left| \vec{X_p}(t) - \vec{A} \cdot \vec{D} \right| \quad (5)$$

The variables used in the Grey Wolf Optimization algorithm include $t$, which represents the current iteration, as well as coefficient vectors represented by $\vec{A}$ and $\vec{A}$. The position vector of the prey is denoted by $X_p$, while the position vector of the grey wolf is represented by $X$. Equations are used to define the $\vec{A}$ and $\vec{A}$ vectors in the algorithm, as described in the source [38].

$$\vec{A} = 2\vec{a} \cdot r_1 - \vec{a}, \quad \vec{C} = 2\vec{r_2} \quad (6)$$

During the Grey Wolf Optimization algorithm, the position of the wolves is updated, where the components of the vector $\vec{a}$ decrease linearly from 2 to 0 over iterations, and $r_1$ and $r_2$ are random vectors between 0 and 1. Through manipulation of the $\vec{A}$ and $\vec{C}$ vectors, the optimal wolf can be identified and repositioned strategically around the prey, as depicted in Fig. 4. In natural circumstances, grey wolves possess the ability to locate their prey, whereas, in mathematical applications, the location of the prey (i.e., the optimal solution) is unknown. Therefore, it is assumed that the $\alpha$, $\beta$, and $\delta$ wolves have a better understanding of the potential location of the prey. The solutions achieved by these three wolves are preserved, while the remaining wolves update their positions based on the saved solutions. This process is mathematically represented by equations that calculate the distances between the alpha, beta, and delta wolves [38].

$$\vec{D_\alpha} = \left| \vec{C_1} \cdot \vec{X_\alpha} - \vec{X} \right|,$$
$$\vec{D_\beta} = \left| \vec{C_2} \cdot \vec{X_\beta} - \vec{X} \right|, \quad (7)$$
$$\vec{D_\delta} = \left| \vec{C_3} \cdot \vec{X_\delta} - \vec{X} \right|$$

Where $\vec{X_\alpha}$, $\vec{X_\beta}$, and $\vec{X_\delta}$ are the positions of $\alpha$, $\beta$, and $\delta$. $\vec{X}$ is the current solution's position.

$$\vec{X_1} = \vec{X_\alpha} - \vec{A_1} \cdot \left( \vec{D_\alpha} \right), \quad (8)$$

$$\vec{X_2} = \vec{X_\beta} - \vec{A_2} \cdot \left(\vec{D_\beta}\right),$$

$$\vec{X_3} = \vec{X_\delta} - \vec{A_3} \cdot \left(\vec{D_\delta}\right)$$

$$\vec{X}(t+1) = \frac{\vec{X_1} + \vec{X_2} + \vec{X_3}}{3}$$

The $C$ Vector is a crucial parameter in the GWO algorithm that controls exploration. It is assigned a random value between 0 and 2 and influences the weight given to the prey based on a wolf's position. The parameter known as $C$ plays a crucial role in determining the level of difficulty in reaching the prey. A value greater than 1 ($C > 1$) makes the task more challenging and increases the distance to the prey, while a value less than 1 ($C < 1$) makes it easier and brings the prey closer. The $C$ vector serves as an effective means to prevent getting stuck in local optima, particularly during the later stages of the algorithm. Another important parameter that facilitates exploration is $A$. Its value is determined by a linearly decreasing parameter that ranges from 2 to 0. When the random values of the $\vec{A}$ vector falls within the interval of [-1,1], a wolf is capable of moving to any position between its current location and the prey's position. If $\left|\vec{A}\right| < 1$, the wolves launch a direct attack on the prey, while values greater than 1 signify that they attack from a distance away from the prey.

## 3.2. Modified grey wolf optimization

To modify the Grey Wolf Optimizer, two simple changes have been proposed. The first change added a new group of wolves called Gamma (γ), in addition to the existing Alpha ($\alpha$), Beta ($\beta$), Delta ($\delta$), and Omega ($\omega$) groups. With this addition, the omega wolves will adjust their positions based on the positions of all four groups of wolves instead of three. The second modification relates to the equation used to determine the step size of the omega wolves, which is presented in the original GWO. In the modified version, an extra equation is included to compute the distance between the alpha, beta, delta, and gamma wolves.

$$\vec{D_\gamma} = \left|\vec{C_4} \cdot \vec{X_\gamma} - \vec{X}\right| \quad (9)$$

The calculation of the positions of alpha, beta, delta, and gamma can be done using the following formulae, where $\vec{X_\gamma}$ represents the position of gamma and $\vec{X}$ is the current solution [37].

$$\vec{D_{avg}} = \frac{\vec{X_1} + \vec{X_2} + \vec{X_3}}{3}$$

$$\vec{X_1} = \vec{X_\alpha} - \vec{A_1} \cdot \left(\vec{D_{avg}}\right), \quad \vec{X_2} = \vec{X_\beta} - \vec{A_2} \cdot \left(\vec{D_{avg}}\right),$$

$$\vec{X_3} = \vec{X_\delta} - \vec{A_3} \cdot \left(\vec{D_{avg}}\right), \quad \vec{X_4} = \vec{X_\delta} - \vec{A_4} \cdot \left(\vec{D_{avg}}\right)$$

$$\vec{X}(t+1) = \frac{\vec{X_1} + \vec{X_2} + \vec{X_3} + \vec{X_4}}{4}$$

(10)

## 3.3. Fitness Dependent Optimizer

The Fitness Dependent Optimizer (FDO) is a recently developed algorithm, introduced by Jaza Abdullah and Tarik Rashid in 2019, that draws inspiration from the reproductive behavior of swarm bees. Unlike GWO, the FDO algorithm adopts a simpler concept and is easier to grasp. It emulates the scouting behavior of bees as they search for a suitable hive among numerous potential options, aiming to identify optimal solutions from a pool of possibilities. The FDO process consists of two main components: the search process, where the related agents make efforts to discover the optimal solution, and the movement process, in which the position of scout bee is updated. Detailed explanations of these two processes will be provided in the following sections [39, 40].

## 3.4. Searching Process of the Scout Bee

The fundamental idea of the Fitness Dependent Optimizer (FDO) is to discover suitable new solutions similar to finding new hives in the bee reproduction process. Scout bees act as search agents to look for new solutions, just like in the GWO algorithm. In the FDO algorithm, a new solution is denoted by the position of a scout bee. During the execution of FDO, the initial positions of artificial scout bees are generated randomly within the search space. Throughout the algorithm, a global best solution is identified. The scout bees utilize a combination of fitness weight mechanisms and random walk to explore and discover new solutions. They persistently search for improved solutions within the predefined boundaries until the end of the process. If a superior solution is discovered, the previous solution is discarded in favor of the new one. However, if no better solution is discovered, the previous solution is retained.

### 3.5. Movement Process of the Scout Bee

The FDO algorithm involves updating the positions of the scout bees to improve the solution. This is done by adding a step size to the current position.

$$X_{i,t+1} = X_{i,t} + pace \quad (11)$$

The search agent at the present moment is symbolized by $i$, while $t$ represents the ongoing iteration. The artificial scout bee or search agent is denoted by $X$, and $pace$ signifies the speed and direction of movement for the scout bee. The pace's speed and direction are influenced by the fitness weight ($f_w$). Nevertheless, the specific direction of the pace is determined randomly. The fitness weight is computed

$$f_w = \left| \frac{x^*_{i,tfitness}}{x_{i,tfitness}} \right| - wf \quad (12)$$

The formula above calculates the weight factor ($wf$) used to control the fitness weight ($f_w$), which determines the movement pace of the scout bees in FDO. $x^*_{i,tfitness}$ and $x_{i,tfitness}$ represent the best global and current solutions for the fitness function, respectively. The weight factor The value of $wf$ can be either 0 or 1, with 1 indicating a high level of convergence and a low chance of coverage. However, if $wf$ is 0, it is ignored and does not affect the equation. It is important to note that setting wf to 0 does not necessarily increase the stability of the search; the fitness function value depends on the problem. To avoid unacceptable cases, the $f_w$ value should be between 0 and 1 and the equation should avoid division by zero, which can occur if the value of $x_{i,tfitness}$ is 0. The following rules, as shown in Eq. (10), should be applied in FDO.

$$\begin{cases} f_w = 1 \text{ or } f_w = 0 \text{ or } x_{i,tfitness} = 0 \text{ or } pace = x_{i,t} \times r \\ f_w > \text{ and } f_w < 1 \begin{cases} r < 0, & pace = (x_{i,t} - x_{i,t}^*) \times f_w \times -1 \\ r \geq 0, & pace = (x_{i,t} - x_{i,t}^*) \times f_w \end{cases} \end{cases} \quad (13)$$

In the given equation, $r$ represents a randomly generated number within the interval of [-1, 1]. The term $x_{i,t}$ denotes the current solution at iteration $t$ for the search agent $i$. On the other hand, $x_{i,t}^*$ refers to the best solution attained globally up until the present moment.

## 4. Implementation, Results, and Discussion

Here, the outcomes achieved through the implemented architecture are analyzed and an evaluation of the training and testing processes are presented. We will provide an overview of the experimental environment and framework employed, as well as showcase the overall performance of the proposed model. Subsequent sections will delve into the specific results obtained from each dataset. The construction and execution of the machine learning classification were conducted using the MATLAB platform. To evaluate the classification model, a dataset was utilized being divided into an 80:20 ratio. This means that 80% of the data was employed for training purposes, while the remaining 20% was reserved for testing.

Tables 1 and 2 present the classification accuracy of each model on the dataset, showcasing the performance during the training and testing phases of the implemented architecture. The datasets were divided into training and testing sets using an 80:20 ratio, comprising a total of 1,941 samples, with 1,553 samples allocated for training and 388 samples for testing. The table provides information on the correct classification rate achieved by each model on each dataset, along with details such as problem dimension, the number of search agents utilized, and the maximum iteration of the search algorithm. All models were tested under identical conditions, employing 10 search agents and a maximum iteration of 50 for each algorithm. From Table 1, it can be observed that all models yielded similar results, but the FDO algorithm exhibited a higher likelihood of achieving more accurate outcomes, attaining 100% accuracy across all experiments. However, it is worth noting that the FDO algorithm took longer to complete compared to the GWO algorithm. Notably, the GWO_MLP model demonstrated the shortest runtime, possibly due to the faster GWO algorithm and the MLP architecture having fewer connections compared to the CMLP.

**Table 1** Proposed models and their correct classification rates.

| Model | Samples | Dimensions | Run time | Training rate | Testing rate |
|---|---|---|---|---|---|
| GWO_MLP | 1941 | 2346 | 139.223s | 92.740 % | 92.87 % |
| MGWO_MLP | 1941 | 2346 | 241.248s | 95.816 % | 96.39 % |
| GWO_CMLP | 1941 | 2482 | 181.149s | 95.217 % | 100.0 % |
| FDO_MLP | 1941 | 2346 | 5637.132s | 100.00 % | 100.0 % |
| FDO_CMLP | 1941 | 2380 | 5837.251s | 99.771 % | 98.65 % |

Table 2 showcases the outcomes of the proposed models, presenting the Mean Square Error (MSE) and the classification rate achieved during both the testing and training phases. The achieved classification rate on this dataset is highly encouraging, particularly considering its substantial size, as it represents the largest dataset utilized in this study. The classification accuracy ranges from 99% to 100%, demonstrating the effectiveness of the models.

**Table 2.** Performance results of the proposed models.

| Model | Train/Test | Pos. case | Correct predicts | Accuracy | Neg. case | Correct predicts | Accuracy | MSE | Rate |
|---|---|---|---|---|---|---|---|---|---|
| GWO_MLP | Tr. | 958 | 876 | 91.379 | 595 | 552 | 92.469 | 0.00268 | 92.740 |
| | Ts. | 311 | 311 | 100.00 | 77 | 72 | 92.215 | 0.00269 | 96.371 |
| MGWO_MLP | Tr. | 958 | 926 | 96.643 | 595 | 556 | 93.266 | 0.00179 | 95.816 |
| | Ts. | 311 | 311 | 100.00 | 77 | 73 | 94.682 | 0.00189 | 97.651 |
| GWO_CMLP | Tr. | 958 | 911 | 95.071 | 595 | 559 | 93.664 | 0.00221 | 95.217 |
| | Ts. | 311 | 311 | 100.00 | 77 | 73 | 93.757 | 0.00243 | 97.171 |
| FDO_MLP | Tr. | 958 | 958 | 100.00 | 595 | 596 | 100.00 | 0.00225 | 100.00 |
| | Ts. | 311 | 311 | 100.00 | 77 | 77 | 100.00 | 0.00244 | 100.00 |
| FDO_CMLP | Tr. | 958 | 947 | 98.764 | 595 | 596 | 100.00 | 0.00199 | 99.771 |
| | Ts. | 311 | 311 | 100.00 | 77 | 77 | 100.00 | 0.00196 | 100.00 |

Additionally, Table 3 provides an evaluation of the confusion matrices for the proposed models on this dataset, offering further insights into their performance. Finally, Fig. 5 illustrates the roc curve results of all the proposed models that underwent testing on the dataset, providing a visual representation of their performance.

**Table 3** Confusion matrices for a dataset using considered methods.

| Model | Sensitivity | Specificity | PPV | NPV | Accuracy |
|---|---|---|---|---|---|
| GWO_MLP | 0.95 | 0.87 | 0.91 | 0.93 | 96.37% |
| MGWO_MLP | 0.98 | 1 | 1 | 0.94 | 97.65% |
| GWO_CMLP | 0.96 | 0.95 | 0.97 | 0.93 | 97.17% |
| FDO_MLP | 0.99 | 1 | 1 | 0.95 | 100% |
| FDO_CMLP | 0.96 | 0.92 | 0.95 | 0.94 | 100% |

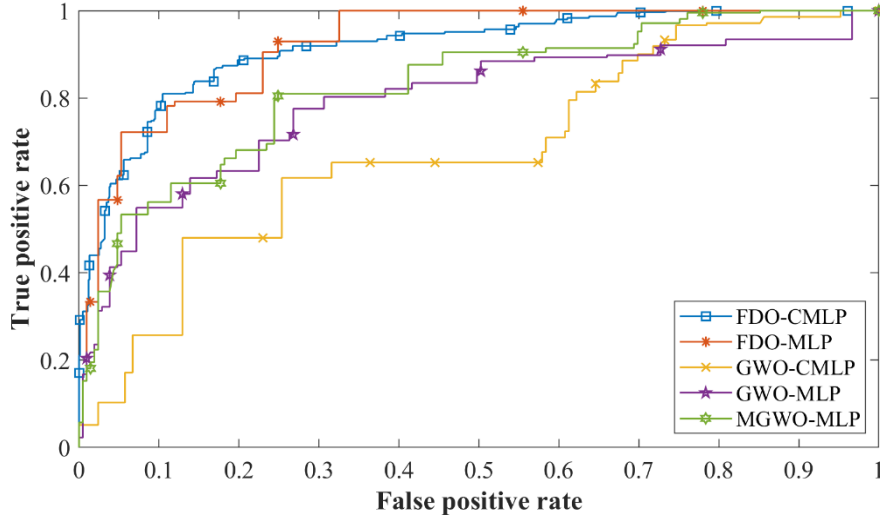

**Fig. 5.** ROC curve results for the dataset.

### 4.1. Conclusions

Early detection of faults in steel plates is crucial for maintaining safety and reliability, and machine learning techniques can accurately predict and diagnose these faults. The investigations conducted in this chapter used five machine-learning techniques to classify steel plates as faulty or non-faulty. In these techniques, the metaheuristic algorithms such as FDO were used to reduce the mean square error.

Based on the results obtained from the experiments, the FDO-based MLP and CMLP models exhibited superior performance compared to the other models, achieving 100% accuracy. Nevertheless, the FDO algorithm exhibits a longer runtime compared to GWO across the dataset. However, in the engineering domain, where accuracy holds paramount importance, the FDO algorithm is considered more favorable. The performance of the other models showed slight variations depending on the specific dataset used. No significant difference was observed between MLP and CMLP models as results varied across experiments. The classification models presented in this study demonstrate impressive precision and offer a promising initial step toward developing a predictive system for identifying infected or high-risk patients, which plays a crucial role in controlling the spread of infections. However, considering the sensitivity of the engineering field, it is crucial to have a highly reliable and efficient system in place. It is important to note that achieving 100% accuracy in classification models is rare and can be influenced by factors such as the number of features used and the ease of classification. Furthermore, the dataset utilized in this research did not include noisy data, which could potentially lead to errors in generalization. Therefore, additional testing is necessary to ensure the reliability of these models. Future work could involve expanding or developing the research by performing additional tests:

- Increasing the database size by adding more features and samples related to the steel plate fault, will enhance the efficiency and comprehensiveness of the prediction process.
- Exploring the creation of alternative neural network models capable of processing diverse data types, including images, audio, and time-series data. This endeavor would result in the development of a more advanced system, characterized by enhanced intelligence and a wide range of advantages, particularly in the engineering field.
- Investigating novel approaches to generate models that are both reliable and advanced. One avenue to accomplish this is by adapting or combining the utilized algorithms with traditional or metaheuristic algorithms to construct innovative models.

## References

1. Ravikumar S, Ramachandran KI, Sugumaran V (2011) Machine learning approach for automated visual inspection of machine components Expert Syst Appl 38 (4): 3260-3266I.
2. He D, Xu K, Zhou P, Zhou D (2019) Surface defect classification of steels with a new semi-supervised learning method. Optics and Lasers in Engineering 117: 40-48.
3. Farahmand-Tabar S, and Ashtari P (2020) Simultaneous size and topology optimization of 3D outrigger-braced tall buildings with inclined belt truss using genetic algorithm. The Structural Design of Tall and Special Buildings 29(13): e1776. https://doi.org/10.1002/tal.1776
4. Tian S, Xu K (2017) An algorithm for surface defect identification of steel plates based on genetic algorithm and extreme learning machine. Metals 7 (8): 311.
5. Ashtari P, Karami R and Farahmand-Tabar S (2021) Optimum geometrical pattern and design of real-size diagrid structures using accelerated fuzzy-genetic algorithm with bilinear membership function. Applied Soft Computing 110: 107646. https://doi.org/10.1016/j.asoc.2021.107646

6. Farahmand-Tabar S, Babaei M (2023) Memory-assisted adaptive multi-verse optimizer and its application in structural shape and size optimization. *Soft Comput*. https://doi.org/10.1007/s00500-023-08349-9
7. Xu K, Xu Y, Zhou P, Wang L (2018) Application of RNAMlet to surface defect identification of steels. Opt Lasers Eng. 105: 110-117.
8. B. Ahn SJ, Ko J (2012) Steel surface defects detection and classification using SIFT and voting strategy. Int J Softw Eng Appl. 6 (2): 161-165.
9. K. Xu, S. Liu, Y. Ai (2015) Application of shearlet transform to classification of surface defects for metals. Image Vis Comput, 35: 23-30.
10. M.P. Paulraj, A.M. Shukry, S. Yaacob, A.H. Adom, R.P. Krishnan (2010) Structural steel plate damage detection using DFT spectral energy and artificial neural network. Signal Processing and Its Applications (CSPA), 6th International Colloquium on, IEEE: 1-6.
11. J.P. Yun, S. Choi, J.W. Kim, S.W. Kim (2009) Automatic detection of cracks in raw steel block using Gabor filter optimized by univariate dynamic encoding algorithm for searches (uDEAS) NDT & E Int, 42 (5): 389-397.
12. A. Landstrom, M.J. Thurley (2012) Morphology-based crack detection for steel slabs. IEEE J Sel Top Signal Process, 6 (7): 866-875.
13. LeCun Y, et al. Gradient-based learning applied to document recognition. In: Proceedings of the IEEE 86.11; 1998. p. 2278–324.
14. Krizhevsky A, Sutskever I, Hinton GE. Imagenet classification with deep convolutional neural networks. In: Advances in neural information processing systems; 2012.p. 1097–105.
15. Simonyan K, Zisserman A, 2014. Very deep convolutional networks for large-scale image recognition. arXiv:1409.1556.
16. Lin M, Chen Q, Yan S, 2013. Network in network. arXiv:1312.4400.
17. Szegedy C, et al. Going deeper with convolutions. In: Proceedings of the IEEE conference on computer vision and pattern recognition; 2015.
18. Szegedy C, et al. Inception-v4, Inception-ResNet and the impact of residual connections on learning AAAI; 2017.
19. Huang G, Liu Z, Van Der Maaten L, Weinberger KQ, et al. Densely connected convolutional networks. In: Girshick R, et al., editors. CVPR. Rich feature hierarchies for accurate object detection and semantic segmentation, 1; 2017. p. 3. Proceedings of the IEEE conference on computer vision and pattern recognition. 2014, July.
20. He K, et al. Spatial pyramid pooling in deep convolutional networks for visual recognition European conference on computer vision Cham. Springer; 2014.
21. Girshick, R. "Fast r-cnn." arXiv:1504.08083 (2015).
22. Ren S, et al. Faster r-cnn: towards real-time object detection with region proposal networks. Adv Neural Inf Process Syst 2015.
23. Redmon J, et al. You only look once: unified, real-time object detection. In: Proceedings of the IEEE conference on computer vision and pattern recognition; 2016.
24. Redmon, J., and A. Farhadi. "YOLO9000: better, faster, stronger." arXiv preprint (2017).

25. Liu W, et al. Ssd: single shot multibox detector European conference on computer vision Cham. Springer; 2016.
26. Ng H-W, et al. (2015) Deep learning for emotion recognition on small datasets using transfer learning. In: Proceedings of the 2015 ACM on international conference on multimodal interaction. ACM.
27. Huynh BQ, Li H, Giger ML. Digital mammographic tumor classification using transfer learning from deep convolutional neural networks. J Med Imaging 3.3 2016:034501.
28. Hoo-Chang S, et al. (2016) Deep convolutional neural networks for computer-aided detection: CNN architectures, dataset characteristics and transfer learning. IEEE Trans Med Imaging 35(5):1285.
29. Li Q, et al. (2014) Medical image classification with convolutional neural network Control Automation Robotics & Vision (ICARCV), 2014 13th International Conference on. IEEE.
30. Masci J, et al. (2011) Stacked convolutional auto-encoders for hierarchical feature extraction International conference on artificial neural networks Berlin, Heidelberg. Springer.
31. dataset provided by Semeion, Research Center of Sciences of Communication, Via Sersale 117, 00128, Rome, Italy.
32. M Buscema, S Terzi, W Tastle, (2010) A New Meta-Classifier,in NAFIPS 2010, Toronto (CANADA), 978-1-4244-7858-6/10 Â©2010 IEEE
33. M Buscema, MetaNet: The Theory of Independent Judges, in Substance Use & Misuse, 33(2), 439-461,1998
34. Steven W (2019) Artificial neural network, in: Advanced Methodologies and Technologies in Artificial Intelligence, Computer Simulation, and Human-Computer Interaction, IGI Global : 40–53.
35. Rashid T (2012).Direct current motor model using RBF, Int. J. Adv. Res. Comput. Sci. Softw. Eng. 2 (9).
36. Mummadisetty BC, Puri A, Sharifahmadian E, Latifi S (2015) A hybrid method for compression of solar radiation data using neural networks. Int. J. Commun., Netw. Syst. Sci. 8 (06): 217.
37. Rashid TA, Abbas DK, Turel YK (2019) A multi hidden recurrent neural network with a modified grey wolf optimizer. *PLOS ONE*, *14*(3), e0213237. https://doi.org/10.1371/journal.pone.0213237
38. Mirjalili S, Mirjalili SM, Lewis A (2014) Grey wolf optimizer, Adv. Eng. Software 69: 46–61.
39. Abdullah JM, Ahmed T (2019) Fitness Dependent Optimizer: Inspired by the Bee Swarming Reproductive Process. in *IEEE Access* 7: 43473-43486. https://doi: 10.1109/ACCESS.2019.2907012.
40. Muhammed DA, Saeed S.A.M, Rashid TA (2020) Improved Fitness-Dependent Optimizer Algorithm. in IEEE Access, 8: 19074-19088. https://doi: 10.1109/ACCESS.2020.2968064.